\documentclass[sigconf]{acmart}

\setcopyright{none}
\renewcommand\footnotetextcopyrightpermission[1]{}

\AtBeginDocument{%
  }

\usepackage{textcomp}
\usepackage{array}
\newcolumntype{P}[1]{>{\centering\arraybackslash}p{#1}}
\usepackage{multirow}
\usepackage{colortbl}
\definecolor{mygray}{gray}{0.90}
\usepackage{pifont}

\newcommand{\xmark}{\raisebox{0.2ex}{\scalebox{0.85}{\ding{55}}}}

\definecolor{rowgray}{gray}{0.96}

\begin{document}

\title{A Unified Tokenization Framework for Pain Recognition using Heterogeneous 3D Modalities}

\author{Stefanos Gkikas}
\orcid{0000-0002-4123-1302}
\affiliation{%
  \institution{Honda Research Institute Japan}
  \city{Wako City}
  \country{Japan}
}
\email{stefanos.gkikas@jp.honda-ri.com}

\author{Christian Arzate Cruz}
\orcid{0000-0003-3433-8379}
\affiliation{%
  \institution{Honda Research Institute Japan}
  \city{Wako City}
  \country{Japan}
}
\email{christian.arzate@jp.honda-ri.com}

\author{Valentina Becchetti}
\orcid{0009-0008-8560-1775}
\affiliation{%
  \institution{Sapienza University of Rome}
  \city{Rome}
  \country{Italy}
}
\email{v.becchetti@diag.uniroma1.it}

\author{Muhammad Umar Khan}
\orcid{0000-0001-6992-6432}
\affiliation{%
  \institution{University of Canberra}
  \city{Canberra}
  \country{Australia}
}
\email{umar.khan@canberra.edu.au}

\author{Alessandro Giuseppi}
\orcid{0000-0001-5503-8506}
\affiliation{%
  \institution{Sapienza University of Rome}
  \city{Rome}
  \country{Italy}
}
\email{giuseppi@diag.uniroma1.it}

\author{Raul Fernandez Rojas}
\orcid{0000-0002-8393-4241}
\affiliation{%
  \institution{University of Canberra}
  \city{Canberra}
  \country{Australia}
}
\email{raul.fernandezrojas@canberra.edu.au}


\renewcommand{\shortauthors}{Gkikas et al.}

\begin{abstract}
Pain is a complex and pervasive phenomenon affecting a large percentage of the population, and accurate assessment is essential for effective clinical management and intervention. Computational pain recognition systems enable continuous monitoring, support clinical decision-making, and help mitigate pain-related distress and functional decline.
This study introduces a unified tokenization framework for heterogeneous 3D modalities in pain recognition that provides a single processing pipeline across behavioral and brain-activity 3D data, without requiring separate architectures for each modality or handcrafted inductive biases. The framework preserves spatial, temporal, and time--frequency structure while mapping diverse inputs into a shared token space.
Extensive experiments show that the proposed approach effectively processes facial videos and fNIRS data in both raw-signal and spectrogram-based representations. On the \textit{AI4Pain} benchmark dataset, the proposed framework achieves state-of-the-art performance while maintaining high computational efficiency and enabling real-time assessment on both GPU and CPU hardware.
\end{abstract}


%\begin{CCSXML}
%<ccs2012>
%      <concept_id>10010405.10010444.10010449</concept_id>
%      <concept_desc>Applied computing~Health informatics</concept_desc>
%      <concept_significance>500</concept_significance>
%      </concept>
%</ccs2012>
%\end{CCSXML}
%
%\ccsdesc[500]{Applied computing~Health informatics}

%\keywords{Pain assessment, deep learning, multimodal, data fusion}


\maketitle
\pagestyle{empty}
\thispagestyle{empty}

\section{Introduction}
Pain serves an important function in the evolution of organisms by providing a warning system about potential damage to the body or disease and aiding in the maintenance of physical integrity \cite{santiago_2022}. As such, pain is subjective and multidimensional, involving noxious, sensory, emotional/affective, and cognitive components \cite{marchand_2024}. Because pain is so common and affects society as a whole, it has been described as a \textit{\textquotedblleft silent public health epidemic\textquotedblright} \cite{katzman_gallagher_2024}. In clinical settings, pain is often called the \textit{\textquotedblleft fifth vital sign,\textquotedblright} underscoring the importance of systematically assessing it during each patient encounter \cite{joel_lucille_1999}. Opioids have historically been one of the most commonly used treatments for pain \cite{kaye_jones_2017}; however, they have also been associated with high rates of abuse and overdose \cite{stampas_pedroza_2020}, and many of these medications cause side effects that limit a person's ability to engage in daily activities and negatively impact the overall quality of their lives \cite{benyamin_trescot_2008}. The complexity of assessing and evaluating pain makes it challenging for researchers and clinicians. The understanding of patients' pain situation and the response to pain management is based on self-reports, which provide limited information regarding the effectiveness of treatments and contribute to the over-prescription and misuse of opioids \cite{kong_chon_2024}. Additionally, when patients are acutely or chronically ill or otherwise unable to express themselves clearly, there may be an increased risk of undertreating their pain \cite{herr_coyne_2019}. 
Pain in adult critical care patients is often poorly managed due to the lack of standardized assessment tools to guide clinical decisions regarding appropriate interventions \cite{meehan_mcrae_1995}. This issue is exacerbated in individuals with cancer-related pain, especially those at end-of-life \cite{snijders_brom_2022}. Finally, chronic pain compromises attention, decreases functional performance, and slows down reaction times \cite{khera_rangasamy_2021}. Li \textit{et al.} \cite{li_lyu_2025} reported that chronic pain alters cognitive control by disrupting expectancy processing and decision-making. Additionally, both chronic and acute/induced pain alter attentional performance depending upon the task requirements and the type of pain being experienced \cite{moore_meints_2019}.

In recent years, modality-agnostic and general-purpose architectures have gained increasing attention in deep learning \cite{jaegle_gimeno_2021}, with growing adoption in biosignal analysis. For example, \textit{BIOT} \cite{yang_westover_2023} analyzes multiple types of biosignals with varying channel configurations using a single model. Unified transformer architectures analyze various forms of visual data, such as facial videos and 2D representations of biosignals \cite{ali_hughes,gkikas_tsiknakis_painvit_2024, gkikas_workload_acii_2026}. Additionally, in \cite{qu_wang_2026}, the authors introduced a symmetry-aware framework that maps heterogeneous physiological signals to a shared latent space for joint affective-state modeling, using separate modality-specific encoders for each signal type. Foundation models for biosignals have also recently emerged, including \textit{PhysioWave} \cite{chen_orlandi_2025} and \textit{Tiny-BioMoE} \cite{gkikas_tiny_2025}, enabling large-scale pre-training across physiological domains. However, despite these advances, current methods typically operate on inputs with shared intrinsic dimensionality, such as 1D waveforms or 2D images. 
In this study, we propose a unified tokenization framework for heterogeneous 3D modalities that preserves spatial, temporal, and time--frequency structure while mapping diverse inputs into a common token space. Facial videos are processed as spatiotemporal volumes, and fNIRS signals as multichannel temporal time series. By folding modality-specific axes and segmenting the token sequence, the framework produces modality-agnostic representations without handcrafted inductive biases while enabling computationally efficient inference on both GPU and CPU hardware.

\section{Related Work}

Over the last decade, there has been a significant expansion in research on the use of computational systems to assess an individual's pain level. While many researchers have shifted toward deep learning-based models to assess a patient's pain, traditional methodologies continue to be used successfully and yield strong results when data are limited \cite{gkikas_phd_thesis_2025}.
Most existing methods rely on video data to capture behavioral manifestations of pain through facial expressions, body motion, and other observable cues, and employ a wide range of learning strategies \cite{huang_dong_2022,bargshady_hirachan_2024,gkikas_tsiknakis_embc,gkikas_reface_acii_2026}. Structured facial representations transfer across affective assessment tasks, as demonstrated by graph-based encodings of action unit dynamics for stress recognition \cite{kassiotis_stressgat_acii_2026}.
Although video-based approaches dominate the literature, a growing body of work has also explored biosignal-driven pain recognition. Prior studies investigated physiological markers derived from electrocardiography \cite{gkikas_chatzaki_2023,gkikas_chatzaki_2022}, electromyography \cite{thiam_bellmann_kestler_2019,patil_patil_2024,pavlidou_tsiknakis_2025}, electrodermal activity (EDA) \cite{ji_zhao_li_2023,phan_iyortsuun_2023,lu_ozek_kamarthi_2023,li_luo_2024,aziz_joseph_2025,gkikas_kyprakis_eda_2025}, respiration rate \cite{gkikas_kyprakis_resp_2025} and brain activity measured via functional near-infrared spectroscopy \cite{rojas_liao_2019,rojas_romero_2021,khan_sousani_2024,rojas_joseph_bargshady_2024,bargshady_aziz_2025, gkikas_arzate_eeite_pain_2026}. Beyond recognition accuracy, explainable modeling has been employed to attribute predictions to specific physiological features and anatomical sites \cite{kyprakis_gkikas_localization_2026}. For broader overviews of pain recognition frameworks and evidence across modalities, we refer to \cite{gkikas_tsiknakis_slr_2023, khan_umar_2025}.

Multimodal pain recognition has also attracted attention, with studies showing that integrating behavioral and physiological information can improve robustness and performance \cite{farmani_bargshady_2025, khan_chetty_2026,gkikas_tsiknakis_thermal_2024}. In \cite{shi_chikhaoui_2022}, the authors fused features extracted from ECG and EDA signals and employed tree-based classifiers. 
Likewise, the authors in \cite{jerritta_murugappan_2022} combined wavelet-based features on ECG and EMG and used a k-nearest-neighbor classifier. Additionally, \cite{liang_luo_2025} demonstrated the benefits of multimodal models by utilizing facial and voice features in pediatric postoperative settings.

Beyond laboratory settings, Badura \textit{et al.} \cite{badura_2021} studied pain during physiotherapy using a multimodal setup that included EDA, EMG, respiration, blood volume pulse (BVP), and hand force measurements. Other approaches focus on jointly modeling physiological and behavioral signals. The authors in \cite{zhi_yu_2019} combined facial videos with ECG, EMG, and EDA by extracting and fusing handcrafted descriptors, while \cite{gkikas_tachos_2024} integrated facial videos with heart rate information within a transformer-based framework for pain intensity estimation. Researchers have also proposed methods that extract physiological information directly from video data and fuse it with the original video features. In \cite{huang_dong_2022}, the authors used 3D CNNs to estimate pseudo heart-rate signals from facial videos and showed that combining these representations improves performance. Similarly, Sun \textit{et al.} \cite{sun_wang_2025} fused features from facial videos with pseudo-biosignals extracted from the same videos, thereby improving pain recognition.

%%%%%%%%%%%%%%%%%%%%%%%%%%%%%%%%%%%%%%%%%%%%%%%%%%%%%%%%%%%%%%

\section{Methodology}

This section presents the proposed unified tokenization framework, along with pre-processing pipelines for facial videos and fNIRS signals. It further describes the augmentation and regularization strategies used for pain recognition.

\subsection{Pre-processing}

The pre-processing stage includes face detection on video frames and the generation of spectrogram representations from the original fNIRS signals. Face detection is performed using the MTCNN detector \cite{zhang_2016}, which applies a cascade of convolutional neural networks to localize facial regions, and the detected faces are resized to $224\times224$ pixels. For the fNIRS modality, the original signals are used without filtering. Both raw 1D fNIRS signals and their spectrogram-based visual representations are considered. Power spectral density (PSD) maps are generated to capture time--frequency energy distributions and resized to $224\times224$ pixels.

\subsection{Architecture}
\label{sec:architecture}

The architecture functions on the unified token representation produced by the proposed tokenization framework, enabling heterogeneous inputs with different dimensionalities and structures to be processed within a single model. A compact set of learned latent vectors interacts with the input token sequence through cross-attention and is iteratively refined via self-attention and feedforward transformations. This approach maintains a fixed latent dimensionality and allows variability in input resolution, sequence length, and modality.
We explored two variants in this study: The first uses a single global latent set that aggregates information from all tokens. The second introduces a structured latent bottleneck by partitioning the token sequence into segments and assigning a latent state to each segment. Both variants share identical tokenization, attention mechanisms, and update rules, differing only in the organization of latent states.
Specifically, for a given input tensor:
\begin{equation}
\mathbf{X} \in \mathbb{R}^{B \times A_1 \times \cdots \times A_D \times C},
\end{equation}
$B$ denotes the batch size, $D$ the number of input axes, $A_1,\ldots,A_D$ the axis sizes, and $C$ the number of channels per input location. The number of axes $D$ remains small (typically $D\in\{1,2\}$) by packing modality-specific factors into the channel dimension.

For 2D inputs, $D=2$ and $A_1 \times A_2 = H \times W$, yielding $\mathbf{X} \in \mathbb{R}^{B \times H \times W \times C}$. Multiple image-like representations per sample (\textit{e.g.}, fNIRS spectrograms or stacked video frames) are concatenated along the channel dimension, with $C$ denoting the total number of stacked channels. For 1D inputs such as biosignal waveforms, $D=1$ and $\mathbf{X} \in \mathbb{R}^{B \times L \times C}$, where $L$ is the temporal length and $C$ denotes the number of signal channels. In most cases, $C=1$, whereas multichannel recordings, such as fNIRS, have $C>1$.
The input axes are flattened into a sequence of $N$ tokens, where $N$ equals the number of spatial or temporal positions (\textit{e.g.}, $N=H\times W$ for $D=2$ and $N=L$ for $D=1$). Each token is a vector derived directly from the input data and positional features. Geometric information is incorporated by encoding the input coordinates using Fourier features. For positions $\mathbf{p}\in[-1,1]^D$, the encoding with $K$ frequency bands and maximum frequency $f_{\max}$ is:
\begin{equation}
\gamma(\mathbf{p}) =
\big[
\sin(\pi s_1 \mathbf{p}),\ \cos(\pi s_1 \mathbf{p}),\ \ldots,\
\sin(\pi s_K \mathbf{p}),\ \cos(\pi s_K \mathbf{p}),\ \mathbf{p}
\big],
\end{equation}
where $\{s_k\}_{k=1}^{K}$ span $[1, f_{\max}/2]$. After flattening, data channels and positional features are concatenated to form the token matrix:
\begin{equation}
\mathbf{T} \in \mathbb{R}^{B \times N \times C'}, \qquad
C' = C + D(2K+1).
\end{equation}

Multi-head attention (MHA) splits queries, keys, and values into $H$ heads, computes scaled dot-product attention independently per head, and concatenates the results:
\begin{equation}
\mathrm{MHA}(\mathbf{Q}, \mathbf{K}, \mathbf{V})
=
\mathrm{Concat}(\mathrm{head}_1,\ldots,\mathrm{head}_H)\mathbf{W}^O,
\end{equation}
where:
\begin{equation}
\mathrm{head}_h
=
\mathrm{softmax}\!\left(
\frac{\mathbf{Q}_h \mathbf{K}_h^\top}{\sqrt{d_h}}
\right)\mathbf{V}_h,
\end{equation}
and $\mathbf{Q}_h=\mathbf{Q}\mathbf{W}^Q_h$, with analogous projections for keys and values.
Layer normalization precedes each attention and feedforward sublayer, and residual connections add the sublayer input to its output.
Cross- and self-attention share the same operator and differ only in the choice of context:
\begin{equation}
\mathrm{MHA}(\mathbf{Q},\mathbf{K},\mathbf{V})
= \mathrm{MHA}(\mathbf{X}\mathbf{W}^Q,\ \mathbf{C}\mathbf{W}^K,\ \mathbf{C}\mathbf{W}^V),
\end{equation}
where the queries always originate from $\mathbf{X}=\mathbf{L}$, and the context equals $\mathbf{C}=\mathbf{T}$ for cross-attention and $\mathbf{C}=\mathbf{L}$ for self-attention.

\subsubsection{\textbf{Global-Latent Variant}}

The global-latent variant maintains a set of $M$ learnable latent vectors:
\begin{equation}
\mathbf{L}^{(0)} \in \mathbb{R}^{M \times d}.
\end{equation}
These latent vectors create a fixed-size latent set that aggregates information from the token sequence through attention. Each layer updates the latent set through attention-based aggregation followed by latent mixing, implemented with pre-normalization, residual connections, and a position-wise feedforward network:
\begin{equation}
\mathbf{L} \leftarrow \mathbf{L} + \mathrm{Attn}(\mathbf{L}, \mathbf{C}),
\end{equation}
where $\mathbf{C}=\mathbf{T}$ for cross-attention and $\mathbf{C}=\mathbf{L}$ for self-attention. 
Here, $\mathrm{Attn}(\mathbf{L},\mathbf{C})$ denotes a pre-normalized multi-head attention block with latent queries $\mathbf{L}$ and context $\mathbf{C}$, followed by a position-wise feedforward transformation. 
The self-attention update is applied a fixed number of times per layer to increase interaction capacity within the latent space. 
After $L$ layers, the resulting latent set $\mathbf{L}^{(L)}$ constitutes the representation used for prediction.

\subsubsection{\textbf{Segment-Latent Variant}}

\begin{figure*}
\begin{center}
\includegraphics[scale=0.65]{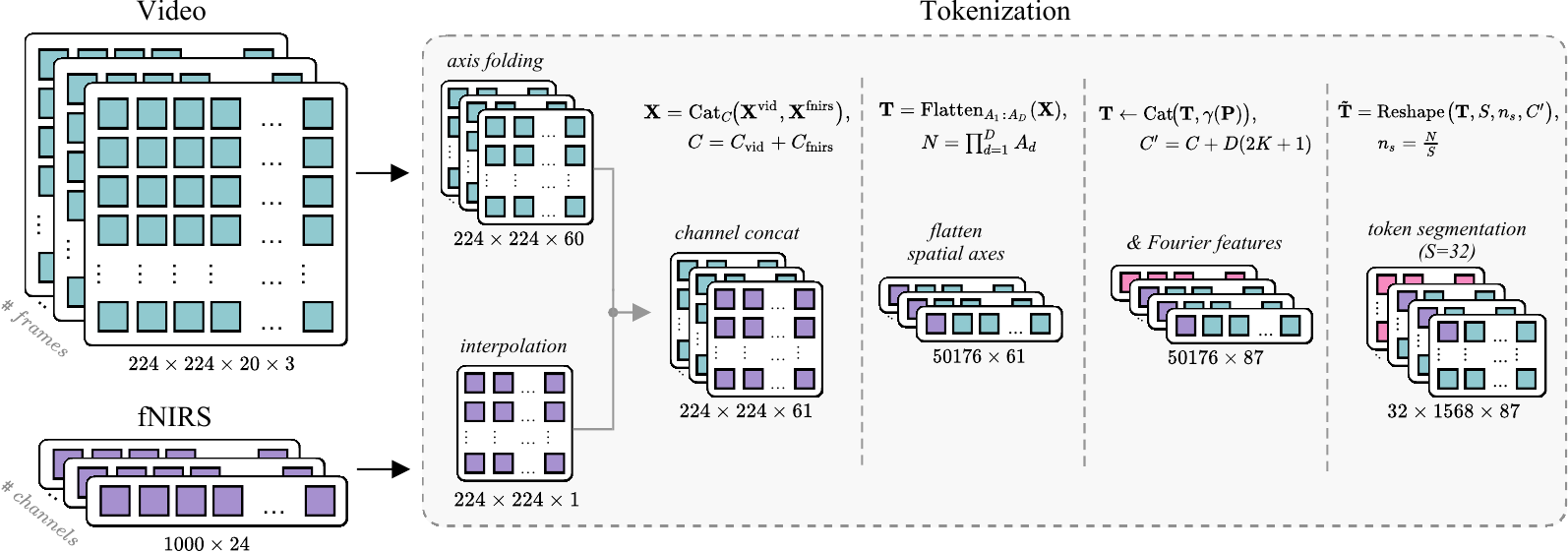} 
\end{center}
\caption{Overview of the proposed tokenization framework, using videos and fNIRS.}
\Description{Overview of the proposed framework for processing video and fNIRS inputs.}
\label{overview}
\end{figure*}

The segment-latent variant introduces a structured latent representation by partitioning the token sequence into $S$ contiguous segments and assigning a latent state to each segment. Tokens are formed as in the global-latent case, $\mathbf{T} \in \mathbb{R}^{B \times N \times C'}$. The sequence is divided into $S$ segments of length $n_s=\lceil N/S \rceil$, padded to $\tilde{N}=Sn_s$, and reshaped as:
\begin{equation}
\tilde{\mathbf{T}} \in \mathbb{R}^{B \times S \times n_s \times C'}.
\end{equation}
The tokens of segment $s$ are denoted $\tilde{\mathbf{T}}_s \in \mathbb{R}^{B \times n_s \times C'}$.
Segment states are instantiated at runtime by replicating a shared initialization vector derived from the global latent parameters:
\begin{equation}
\boldsymbol{\ell}_{\mathrm{init}} = \frac{1}{M}\sum_{m=1}^{M}\boldsymbol{\ell}_m \in \mathbb{R}^{d},
\end{equation}
across segments and batch elements. These segment states are not independent learnable parameters; segment-specific behavior emerges through segment-local attention updates.
Each segment state aggregates information only from its corresponding token subset via segment-local cross-attention:
\begin{equation}
\mathbf{e}^{(\ell)}_s
=
\mathbf{e}^{(\ell-1)}_s
+
\mathrm{Attn}\!\big(\mathbf{e}^{(\ell-1)}_s,\ \tilde{\mathbf{T}}_s\big),
\end{equation}
where $\mathbf{e}^{(\ell)}_s \in \mathbb{R}^{B \times d}$. Stacking all segment states yields $\mathbf{E}^{(\ell)} \in \mathbb{R}^{B \times S \times d}$. Information exchange across segments is then performed through self-attention over the segment-state matrix:
\begin{equation}
\mathbf{E}^{(\ell)} \leftarrow
\mathbf{E}^{(\ell)} +
\mathrm{Attn}\ \!\big(\mathbf{E}^{(\ell)},\ \mathbf{E}^{(\ell)}\big),
\end{equation}
which is repeated $R$ times per layer. Segment-local cross-attention is executed in parallel by packing segments into the batch dimension, without altering the underlying computation. After the final layer, the segment states $\mathbf{E}^{(L)}$ form the representation used for prediction.
Figure \ref{overview} illustrates the \textit{Segment-Latent} tokenization variant applied to facial video and fNIRS modalities in a multimodal setting.
Both variants reuse the same attention and feedforward weights; they differ in the latent structure and attention pattern (global $M$ latents attending to all tokens versus $S$ segment latents attending locally, followed by self-attention across segments). 
The model hyperparameters used in all experiments are summarized in Table \ref{tab:architecture_details}.

\begin{table}
\small
\caption{Architectural details.}
\label{tab:architecture_details}
\begin{tabular}{l c}
\toprule
\textbf{Hyperparameter} & \textbf{Value} \\
\midrule
\midrule
Depth ($L$) & 4 \\
Number of latents ($M$) & 32 \\
Latent dimension ($d$) & 128 \\
Cross-attention heads & 1 \\
Latent self-attention heads & 8 \\
Cross-attention head dimension & 64 \\
Latent attention head dimension & 64 \\
Self-attention blocks per cross ($R$) & 8 \\
\bottomrule
\end{tabular}
\end{table}

%%%%%%%%%%%%%%%%%%%%%%%%%%%%%%%%%%%%%%%%%%%%%%%%%%%%%%%%%%%%%%%%%%%%%%%%%%%%%%%%%

\subsection{Augmentation Methods \& Regularization}
Several augmentation and regularization methods were applied across all experiments. For image-based inputs (video frames and fNIRS spectrograms), \textit{TrivialAugment} \cite{trivialAugment}, \textit{AugMix} \cite{augmix}, center cropping, additive noise, and spatial \textit{Masking} were employed. The same randomly sampled transform was consistently applied across frames or across stacked channels within each sample.
For 1D fNIRS waveforms, temporal length was standardized via truncation/padding, followed by polarity inversion, additive noise, and temporal \textit{Masking} of contiguous segments, again applied consistently across channels. Regularization included \textit{Label Smoothing}, \textit{Attention Dropout}, and \textit{Feedforward Dropout}. Table \ref{table:augm_regul} summarizes all settings.

%%%%%%%%%%%%%%%%%%%%%%%%%%%%%%%%%%%%%%%%%%%%%%%%%%%%%%%%%%%%%%%%%%%%%%%%%%%%%%
\begin{table*}
\footnotesize
\caption{Augmentation and regularization details for each modality.}
\label{table:augm_regul}
\centering

\begin{tabular}{P{1.3cm} P{0.9cm} P{0.9cm} P{1.3cm} P{0.9cm} P{1.3cm} P{1.1cm} P{1.2cm} P{1.85cm} P{0.3cm} P{1.3cm} P{1.2cm}}
\toprule
\multirow{2}{*}{\shortstack{Modality}} &
\multicolumn{8}{c}{Augmentation} &
\multicolumn{3}{c}{Regularization} \\
\cmidrule(lr){2-9}\cmidrule(lr){10-12}

& \textit{Trivial}
& \textit{AugMix}
& \textit{CenterCrop}
& \textit{Polarity}
& \textit{Noise}
& \textit{Masking-1}
& \textit{Masking-2}
& \textit{Masking-3}
& \textit{LS}
& \textit{Att-DropOut}
& \textit{FF-DropOut} \\

\midrule
\midrule

Video/PSD
& 0.20-0.50
& 0.20-0.50
& 0.20-0.50\textbar 200
& ---
& 0.20-0.50\textbar 100
& 0.10-0.80\textbar 3
& 0.20-0.50\textbar 20
& ---
& 0.10
& 0.10
& 0.10 \\

\rowcolor{rowgray}
fNIRS-Wave
& ---
& ---
& ---
& 0.20-0.50
& 0.20-0.50\textbar 100
& ---
& ---
& 0.40-0.50\textbar 0.15-0.30
& 0.10
& 0.10
& 0.10 \\

\bottomrule
\end{tabular}

\vspace{2pt}

\parbox{\textwidth}{%
\scriptsize
\textit{LS}: label smoothing coefficient.
\textit{Att-DropOut}/\textit{FF-DropOut}: dropout probability in attention/feed-forward sublayers.
\textit{Masking-1/2}: Cutout on images (video and fNIRS-PSD), using $32\times32$ squares; the value after \textbar\ indicates the number of blocks.
\textit{Masking-3}: temporal masking on fNIRS waveforms; the value after \textbar\ indicates the masked fraction range.
\textit{Notes}: $x_1$--$x_2$ means that $p\sim\mathcal{U}(x_1,x_2)$ is sampled per sample and the transform is applied with probability $p$.
$x_1$--$x_2\textbar y$ or $x_1$--$x_2\textbar y_1$--$y_2$ additionally specifies the corresponding transform parameter or parameter range.
}

\end{table*}

\section{Experimental Evaluation \& Results}
This study utilizes the AI4Pain dataset \cite{ai4pain_2024,rojas_hirachan_2023}, which contains synchronized facial videos and functional near-infrared spectroscopy (fNIRS) recordings from $65$ individuals. Data were collected at the Human--Machine Interface Laboratory at the University of Canberra, Australia. Of the $65$ subjects, $41$ were used for training, $12$ for validation, and $12$ for testing. Transcutaneous electrical nerve stimulation electrodes were applied to the inner forearm and the dorsum of the subject's right hand in order to elicit controlled pain responses. The pain threshold is the minimum stimulus intensity that is perceived as painful (low pain), while pain tolerance is the maximum stimulus intensity that an individual can endure before it becomes intolerable (high pain). For the fNIRS modality, $24$ oxygenated hemoglobin (HbO) channels were utilized, and facial video sequences were sampled at $30$ frames per second.

All experiments follow a three-class setting (No Pain, Low Pain, High Pain) on the validation set, with performance reported using macro-averaged accuracy, precision, and F1 score. In addition, we present comparisons with prior work on the testing set in Section \ref{comparison_testing}.

\subsection{Global-Latent Variant}
Table \ref{table:v1_results} summarizes the performance and computing characteristics of the \textit{Global-Latent} version of the tokenization method using facial video and fNIRS modalities. Video inputs were analyzed at various temporal resolutions using temporal subsampling. We analyze how differences in input resolution (and thus in the dimensionality of the tokenized representation) will influence the size of cross-attentions in each layer of our multi-layer attention network; as a result, we observe variability in the number of parameters and GFLOPs across configurations.

As we decrease the number of sample frames from $300$ to $10$, both model size and computational cost decrease dramatically, resulting in a significant reduction in inference latency and increased throughput on both GPUs and CPUs. 
Regardless of the substantial reduction in our models' compute cost, the accuracy does remain relatively similar with respect to the number of frames. We achieve an accuracy of $44.81\%$ using $100$ sampled frames. Several other configurations with fewer samples achieve similar performance, indicating that very high densities of sampled frames are not necessary to perform well in pain classification. When we combine video-based results, we find an average accuracy of $42.39\%$ across a variety of configurations. Furthermore, using $150$ sampled frames yields the highest precision ($57.28\%$), while using $100$ sampled frames achieves the best balance between accuracy and F1-score. 
This suggests that different temporal resolutions affect the performance metrics in different ways.

For the evaluation of the fNIRS modality, we use two representations: first, we evaluate the raw waveform data; second, we evaluate the spectrogram-based PSD representation using all $24$ available channels. The PSD representation results in higher overall performance, achieving an accuracy of $49.26\%$ and outperforming the raw waveform input. This indicates that the time--frequency structure within the PSD representation provides richer information about pain. Since the PSD representation is a higher-dimensional representation than the raw waveform input, it requires significantly more computational resources to process. However, both fNIRS configurations require fewer computational resources than most video-based configurations. Overall across both modalities, the \textit{Global-Latent} version of the proposed method achieved an average accuracy of $44.49\%$, while fNIRS achieved $46.60\%$.

%%%%%%%%%%%%%%%%%%%%%%%%%%%%%%%%%%%%%%%%%%%%%%%%%
\begin{table*}
\footnotesize
\caption{Performance and computational/inference cost of the \textit{Global-Latent} variant.}
\label{table:v1_results}
\centering

\begin{tabular}{P{1.3cm} P{1.1cm} P{1.1cm} P{0.9cm} P{2.0cm} P{1.7cm} P{2.0cm} P{1.7cm} P{0.8cm} P{0.7cm} P{0.6cm}}
\toprule

\multirow{2}{*}{\shortstack{Modality}} &
\multirow{2}{*}{\shortstack{\#Channels}} &
\multicolumn{2}{c}{Computational Cost} &
\multicolumn{4}{c}{Inference Cost} &
\multicolumn{3}{c}{Performance} \\

\cmidrule(lr){3-4}\cmidrule(lr){5-8}\cmidrule(lr){9-11}
& & Params(M) & GFLOPs &
Latency (ms) GPU$\downarrow$ &
Samples/s GPU$\uparrow$ &
Latency (ms) CPU$\downarrow$ &
Samples/s CPU$\uparrow$ &
Accuracy & Precision & F1 \\

\midrule
\midrule

Video &300 &98.36 &121.43 &91.81 &10.89 &283.55 &3.53  &43.61 &41.09 &39.42 \\\rowcolor{rowgray}
Video &150 &53.09 &63.63  &46.16 &21.67 &161.61 &6.19  &\underline{43.89} &\textbf{57.28} &42.11 \\
Video &100 &37.99 &44.36  &29.73 &33.63 &125.46 &7.97  &\textbf{44.81} &47.83 &\textbf{43.04} \\\rowcolor{rowgray}
Video &75  &30.45 &34.73  &20.25 &49.39 &98.08  &10.20 &43.52 &36.00 &39.40 \\
Video &60  &25.92 &28.95  &16.37 &61.09 &79.28  &12.61 &42.08 &49.05 &30.86 \\\rowcolor{rowgray}
Video &50  &22.90 &25.09  &14.19 &70.45 &66.13  &15.12 &40.19 &\underline{50.51} &29.25 \\
Video &43  &20.79 &22.40  &13.65 &73.28 &61.11  &16.36 &42.21 &43.69 &\underline{42.89} \\\rowcolor{rowgray}
Video &38  &19.28 &20.47  &14.16 &70.61 &61.25  &16.33 &42.31 &26.06 &28.50 \\
Video &33  &18.07 &18.93  &14.33 &69.78 &66.10  &15.13 &43.06 &43.21 &42.65 \\\rowcolor{rowgray}
Video &30  &16.86 &17.39  &13.74 &72.78 &53.15  &17.81 &41.62 &42.85 &40.01 \\
Video &20  &13.85 &13.53  &14.59 &68.54 &45.37  &22.04 &42.04 &45.70 &32.29 \\\rowcolor{rowgray}
Video &15  &12.34 &11.61  &14.35 &69.68 &51.00  &19.61 &40.79 &41.39 &36.90 \\
Video &10  &10.83 &9.68   &13.64 &73.30 &38.32  &26.09 &40.88 &42.00 &33.26 \\\rowcolor{rowgray}

\midrule

fNIRS-Wave &24 &7.81  &2.58  &14.26 &70.14 &22.44 &44.56 &43.94 &42.94 &39.34 \\
fNIRS-PSD  &24 &15.05 &15.07 &14.36 &69.64 &56.14 &17.81 &\textbf{49.26} &\textbf{47.91} &\textbf{39.90} \\

\bottomrule
\end{tabular}

\vspace{2pt}

\parbox{\textwidth}{%
\scriptsize
\emph{Notes:} For Video, the effective channel count equals the number of sampled frames $\times 3$ (RGB); the column reports the frame count for brevity. For fNIRS, each signal channel contributes one scalar per time step; the column reports the number of HbO signal channels (24). Bold indicates the highest score within each modality, while underlining indicates the second-highest.
}

\end{table*}

\subsection{Segment-Latent Variant}

The performance and computational characteristics of the \textit{Segment-Latent} tokenization variant, as a function of the number of segments, are shown in Tables \ref{table:v3_2_segments}--\ref{table:v3_64_segments}. 
As in the \textit{Global-Latent} version, different sampling rates were applied to the videos, and we used all $24$ available channels for each fNIRS representation. 
We note that increasing the number of segments also increases the number of segment-level self-attentions that need to be computed; however, it does not affect the size of the tokens being processed. Therefore, as the number of segments increases, the influence on GFLOPs and latency is small.

As with the \textit{Global-Latent} version, we observe stable performance across all video configurations with the \textit{Segment-Latent} version. The highest video performance occurs at intermediate frame resolution. Specifically, the highest video classification accuracies are achieved approximately in the range of $38$--$100$ sampled frames for most segment numbers, reaching $45.79\%$ with $2$ segments, $45.74\%$ with $4$ segments, $45.14\%$ with $8$ segments, $47.18\%$ with $16$ segments, $48.94\%$ with $32$ segments, and $46.57\%$ with $64$ segments.
The precision and F1 score exhibit complementary behaviors. Some temporal resolutions tend to produce higher precision (\textit{e.g.}, $150$ frames for $2$ segments and $15$ frames for $4$ segments) than others, which tend to strike a better balance between accuracy and F1. When averaged across all of the video configurations, we observe mean accuracies of $43.28\%$, $42.90\%$, $42.60\%$, $43.10\%$, $44.05\%$, and $43.46\%$ for $2$, $4$, $8$, $16$, $32$, and $64$ segments, respectively, for the \textit{Segment-Latent} model. This indicates consistent performance across various segment resolutions.

\begin{figure}
\begin{center}
\includegraphics[scale=0.58]{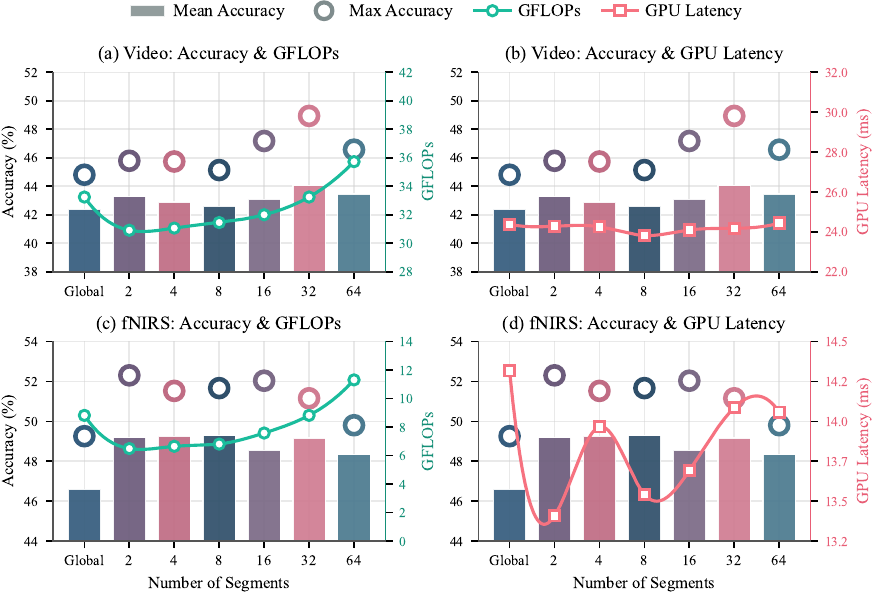} 
\end{center}
\caption{Comparison of the \textit{Global-Latent} and \textit{Segment-Latent} variants in terms of accuracy, GFLOPs, and GPU latency for the video and fNIRS modalities.}
\Description{Comparison of the \textit{Global-Latent} and \textit{Segment-Latent} variants in terms of accuracy, GFLOPs, and GPU latency for the video and fNIRS modalities.}
\label{accuracy_cost}
\end{figure}

The results for the fNIRS modality are similar. The spectrogram-based PSD representation consistently outperforms the raw waveform input across all segment configurations. 
We found the best performance at $52.31\%$ with $2$ segments.
Similarly, regardless of the number of segments used, the waveform inputs performed similarly well with a slight decrease in accuracy compared to PSD inputs. Across fNIRS configurations, we observed mean accuracies ranging from $48.33\%$ to $49.29\%$. 
Thus, the number of segments has little effect on fNIRS performance.
When we combine the results from both modalities and compute their average accuracy, we observe very high consistency and low variance in the resulting accuracies. Specifically, we observed average accuracies ranging from $45.82\%$ to $46.61\%$. Finally, we found that the maximum average value was attained at $32$ segments ($46.61\%$). Compared to the \textit{Global-Latent} variant, the \textit{Segment-Latent} tokenization consistently achieves higher performance while requiring lower computational and inference cost, indicating a more effective latent organization. The trends in performance relative to GFLOPs and GPU latency for both tokenization variants are illustrated in Fig. \ref{accuracy_cost}.

%%%%%%%%%%%%%%%%%%%%%%%%%%%%%%%%%%%

%%%%%%%%%%%%%%%%%%%%%%%%%%%%%%%%%%%%%%%%%%%%%%%%%
\begin{table*}
\footnotesize
\caption{Performance and computational/inference cost of the \textit{Segment-Latent} variant with 2 segments.}
\label{table:v3_2_segments}
\centering

\begin{tabular}{P{1.3cm} P{0.50cm} P{0.55cm} P{1.1cm} P{0.9cm} P{2.0cm} P{1.7cm} P{2.0cm} P{1.7cm} P{0.7cm} P{0.6cm} P{0.4cm}}
\toprule

\multirow{2}{*}{\shortstack{Modality}} &
\multirow{2}{*}{\shortstack{\#Chan.}} &
\multirow{2}{*}{\shortstack{\#Segm.}} &
\multicolumn{2}{c}{Computational Cost} &
\multicolumn{4}{c}{Inference Cost} &
\multicolumn{3}{c}{Performance} \\

\cmidrule(lr){4-5}\cmidrule(lr){6-9}\cmidrule(lr){10-12}

& & &
Params(M) &
GFLOPs &
Latency (ms) GPU$\downarrow$ &
Samples/s GPU$\uparrow$ &
Latency (ms) CPU$\downarrow$ &
Samples/s CPU$\uparrow$ &
Accuracy &
Precision &
F1 \\

\midrule
\midrule

Video &300 &2 &98.36 &119.10 &92.79 &10.78 &355.59 &2.81  &42.87 &41.40 &41.06 \\\rowcolor{rowgray}
Video &150 &2 &53.09 &61.30  &48.98 &20.84 &180.05 &5.55  &43.94 &\textbf{49.97} &32.86 \\
Video &100 &2 &37.99 &42.03  &28.23 &35.42 &124.92 &8.01  &45.05 &42.94 &37.85 \\\rowcolor{rowgray}
Video &75  &2 &30.45 &32.40  &19.98 &50.06 &98.18  &10.19 &41.39 &41.25 &38.64 \\
Video &60  &2 &25.92 &26.62  &17.06 &58.60 &85.58  &11.69 &43.47 &43.27 &\underline{41.69} \\\rowcolor{rowgray}
Video &50  &2 &22.90 &22.76  &13.34 &74.94 &78.65  &12.71 &43.01 &42.76 &41.30 \\
Video &43  &2 &20.79 &20.07  &14.06 &71.10 &71.59  &13.97 &40.23 &\underline{48.98} &28.49 \\\rowcolor{rowgray}
Video &38  &2 &19.28 &18.14  &13.53 &73.92 &66.94  &14.94 &\textbf{45.79} &48.01 &40.14 \\
Video &33  &2 &18.07 &16.60  &13.66 &73.21 &69.22  &14.45 &41.11 &43.31 &33.04 \\\rowcolor{rowgray}
Video &30  &2 &16.86 &15.06  &13.32 &75.10 &53.63  &18.65 &42.04 &24.94 &27.50 \\
Video &20  &2 &13.85 &11.20  &13.23 &75.57 &48.64  &20.56 &\underline{45.28} &45.27 &40.40 \\\rowcolor{rowgray}
Video &15  &2 &12.34 &9.28   &13.50 &74.07 &39.49  &25.32 &43.80 &43.87 &\textbf{42.62} \\
Video &10  &2 &10.83 &7.35   &14.02 &71.32 &34.81  &28.73 &44.68 &46.84 &36.50 \\\rowcolor{rowgray}

\midrule

fNIRS-Wave &24 &2 &7.81  &0.25  &13.33 &74.99 &11.33 &88.27 &\underline{46.06} &\underline{42.63} &\underline{43.20} \\
fNIRS-PSD  &24 &2 &15.05 &12.74 &13.40 &74.62 &51.03 &19.59 &\textbf{52.31} &\textbf{46.66} &\textbf{47.76} \\

\bottomrule
\end{tabular}

\vspace{2pt}

\parbox{\textwidth}{%
\scriptsize
\textit{Chan.}: Channels.
\textit{Segm.}: Segments.
}

\end{table*}
%%%%%%%%%%%%%%%%%%%%%%%%%%%%%%%%%%%%%%%%%%%%%%%%%
%%%%%%%%%%%%%%%%%%%%%%%%%%%%%%%%%%%%%%%%%%%%%%%%%
\begin{table*}
\footnotesize
\caption{Performance and computational/inference cost of the \textit{Segment-Latent} variant with 4 segments.}
\label{table:v3_4_segments}
\centering

\begin{tabular}{P{1.3cm} P{0.50cm} P{0.55cm} P{1.1cm} P{0.9cm} P{2.0cm} P{1.7cm} P{2.0cm} P{1.7cm} P{0.7cm} P{0.6cm} P{0.4cm}}
\toprule

\multirow{2}{*}{\shortstack{Modality}} &
\multirow{2}{*}{\shortstack{\#Chan.}} &
\multirow{2}{*}{\shortstack{\#Segm.}} &
\multicolumn{2}{c}{Computational Cost} &
\multicolumn{4}{c}{Inference Cost} &
\multicolumn{3}{c}{Performance} \\

\cmidrule(lr){4-5}\cmidrule(lr){6-9}\cmidrule(lr){10-12}
& & & Params(M) & GFLOPs &
Latency (ms) GPU$\downarrow$ &
Samples/s GPU$\uparrow$ &
Latency (ms) CPU$\downarrow$ &
Samples/s CPU$\uparrow$ &
Accuracy & Precision & F1 \\

\midrule
\midrule

Video &300 &4 &98.36 &119.26 &90.73 &11.02 &296.58 &3.37  &41.99 &25.24 &28.31 \\\rowcolor{rowgray}
Video &150 &4 &53.09 &61.45  &47.12 &21.22 &160.69 &6.22  &43.94 &44.28 &\underline{42.28} \\
Video &100 &4 &37.99 &42.19  &27.81 &35.96 &123.87 &8.07  &\textbf{45.74} &\underline{45.92} &38.17 \\\rowcolor{rowgray}
Video &75  &4 &30.45 &32.55  &22.90 &43.66 &91.88  &10.88 &44.17 &44.43 &41.33 \\
Video &60  &4 &25.92 &26.77  &16.70 &59.88 &81.69  &12.24 &42.36 &43.12 &\textbf{42.29} \\\rowcolor{rowgray}
Video &50  &4 &22.90 &22.92  &13.84 &72.25 &62.02  &16.12 &43.01 &42.18 &41.10 \\
Video &43  &4 &20.79 &20.22  &13.57 &73.68 &61.43  &16.28 &41.39 &41.82 &40.08 \\\rowcolor{rowgray}
Video &38  &4 &19.28 &18.29  &14.39 &69.49 &54.74  &18.27 &\underline{45.19} &45.74 &34.12 \\
Video &33  &4 &18.07 &16.75  &13.85 &72.22 &56.28  &17.77 &39.12 &25.73 &22.52 \\\rowcolor{rowgray}
Video &30  &4 &16.86 &15.21  &13.48 &74.17 &48.97  &20.42 &43.06 &24.75 &30.24 \\
Video &20  &4 &13.85 &11.36  &13.52 &73.98 &40.63  &24.61 &43.15 &42.58 &40.28 \\\rowcolor{rowgray}
Video &15  &4 &12.34 &9.43   &13.90 &71.95 &38.38  &26.06 &41.76 &\textbf{52.34} &29.56 \\
Video &10  &4 &10.83 &7.50   &13.54 &73.86 &33.51  &29.84 &42.78 &43.40 &40.71 \\\rowcolor{rowgray}

\midrule

fNIRS-Wave &24 &4 &7.81  &0.41  &14.31 &69.90 &11.80 &84.75 &46.99 &44.32 &43.59 \\
fNIRS-PSD  &24 &4 &15.05 &12.90 &13.58 &73.63 &45.02 &22.21 &\textbf{51.53} &\textbf{46.12} &\textbf{47.11} \\

\bottomrule
\end{tabular}

\end{table*}
%%%%%%%%%%%%%%%%%%%%%%%%%%%%%%%%%%%%%%%%%%%%%%%%%

%%%%%%%%%%%%%%%%%%%%%%%%%%%%%%%%%%%%%%%%%%%%%%%%%
\begin{table*}[h]
\footnotesize
\caption{Performance and computational/inference cost of the \textit{Segment-Latent} variant with 8 segments.}
\label{table:v3_8_segments}
\centering

\begin{tabular}{P{1.3cm} P{0.50cm} P{0.55cm} P{1.1cm} P{0.9cm} P{2.0cm} P{1.7cm} P{2.0cm} P{1.7cm} P{0.7cm} P{0.6cm} P{0.4cm}}
\toprule

\multirow{2}{*}{\shortstack{Modality}} &
\multirow{2}{*}{\shortstack{\#Chan.}} &
\multirow{2}{*}{\shortstack{\#Segm.}} &
\multicolumn{2}{c}{Computational Cost} &
\multicolumn{4}{c}{Inference Cost} &
\multicolumn{3}{c}{Performance} \\

\cmidrule(lr){4-5}\cmidrule(lr){6-9}\cmidrule(lr){10-12}
& & & Params(M) & GFLOPs &
Latency (ms) GPU$\downarrow$ &
Samples/s GPU$\uparrow$ &
Latency (ms) CPU$\downarrow$ &
Samples/s CPU$\uparrow$ &
Accuracy & Precision & F1 \\

\midrule
\midrule

Video &300 &8 &98.36 &119.57 &90.10 &11.10 &342.15 &2.92  &43.56 &43.75 &\underline{43.17} \\\rowcolor{rowgray}
Video &150 &8 &53.09 &61.76  &45.94 &21.33 &188.64 &5.30  &40.79 &42.33 &37.14 \\
Video &100 &8 &37.99 &42.50  &27.67 &36.14 &127.59 &7.84  &41.94 &43.98 &41.21 \\\rowcolor{rowgray}
Video &75  &8 &30.45 &32.86  &20.48 &48.82 &105.21 &9.51  &39.17 &41.85 &36.91 \\
Video &60  &8 &25.92 &28.08  &16.66 &60.02 &83.38  &11.99 &41.90 &40.48 &38.78 \\\rowcolor{rowgray}
Video &50  &8 &22.90 &23.23  &13.43 &74.45 &75.28  &13.28 &\underline{45.09} &44.65 &38.51 \\
Video &43  &8 &20.79 &20.53  &14.18 &70.50 &58.52  &17.09 &41.20 &41.26 &40.43 \\\rowcolor{rowgray}
Video &38  &8 &19.28 &18.60  &13.61 &73.45 &54.61  &18.31 &41.62 &41.23 &40.60 \\
Video &33  &8 &18.07 &17.06  &13.37 &74.82 &64.62  &15.48 &43.94 &43.91 &40.80 \\\rowcolor{rowgray}
Video &30  &8 &16.86 &15.52  &13.28 &75.29 &54.83  &18.24 &41.62 &24.57 &27.61 \\
Video &20  &8 &13.85 &11.67  &13.90 &71.92 &44.39  &22.53 &\textbf{45.14} &\underline{45.92} &\textbf{44.23} \\\rowcolor{rowgray}
Video &15  &8 &12.34 &9.74   &13.43 &74.46 &38.29  &26.12 &45.00 &44.99 &42.33 \\
Video &10  &8 &10.83 &7.81   &13.60 &73.55 &33.42  &29.92 &42.87 &\textbf{48.80} &34.77 \\\rowcolor{rowgray}

\midrule

fNIRS-Wave &24 &8 &7.81  &0.41  &13.62 &73.39 &13.80 &72.48 &46.90 &43.74 &42.96 \\
fNIRS-PSD  &24 &8 &15.05 &13.21 &13.39 &74.70 &45.98 &21.75 &\textbf{51.67} &\textbf{49.58} &\textbf{43.06} \\

\bottomrule
\end{tabular}

\end{table*}
%%%%%%%%%%%%%%%%%%%%%%%%%%%%%%%%%%%%%%%%%%%%%%%%%

%%%%%%%%%%%%%%%%%%%%%%%%%%%%%%%%%%%%%%%%%%%%%%%%%
\begin{table*}
\footnotesize
\caption{Performance and computational/inference cost of the \textit{Segment-Latent} variant with 16 segments.}
\label{table:v3_16_segments}
\centering

\begin{tabular}{P{1.3cm} P{0.50cm} P{0.55cm} P{1.1cm} P{0.9cm} P{2.0cm} P{1.7cm} P{2.0cm} P{1.7cm} P{0.7cm} P{0.6cm} P{0.4cm}}
\toprule

\multirow{2}{*}{\shortstack{Modality}} &
\multirow{2}{*}{\shortstack{\#Chan.}} &
\multirow{2}{*}{\shortstack{\#Segm.}} &
\multicolumn{2}{c}{Computational Cost} &
\multicolumn{4}{c}{Inference Cost} &
\multicolumn{3}{c}{Performance} \\

\cmidrule(lr){4-5}\cmidrule(lr){6-9}\cmidrule(lr){10-12}
& & & Params(M) & GFLOPs &
Latency (ms) GPU$\downarrow$ &
Samples/s GPU$\uparrow$ &
Latency (ms) CPU$\downarrow$ &
Samples/s CPU$\uparrow$ &
Accuracy & Precision & F1 \\

\midrule
\midrule

Video &300 &16 &98.36 &120.19 &91.27 &10.96 &351.78 &2.84  &46.30 &\textbf{49.96} &34.61 \\\rowcolor{rowgray}
Video &150 &16 &53.09 &62.39  &47.63 &22.14 &181.53 &5.51  &41.34 &39.27 &\textbf{47.35} \\
Video &100 &16 &37.99 &43.12  &28.15 &35.53 &126.57 &7.90  &\underline{46.39} &44.20 &\underline{43.80} \\\rowcolor{rowgray}
Video &75  &16 &30.45 &33.48  &20.29 &49.29 &98.65  &10.14 &41.39 &39.55 &38.11 \\
Video &60  &16 &25.92 &27.70  &16.68 &59.94 &87.08  &11.48 &42.18 &24.83 &27.87 \\\rowcolor{rowgray}
Video &50  &16 &22.90 &23.85  &13.60 &73.54 &76.37  &13.09 &\textbf{47.18} &\underline{45.78} &40.70 \\
Video &43  &16 &20.79 &21.15  &14.23 &70.30 &64.57  &15.49 &41.62 &40.24 &38.21 \\\rowcolor{rowgray}
Video &38  &16 &19.28 &19.23  &13.48 &74.19 &58.53  &17.09 &45.28 &44.02 &41.57 \\
Video &33  &16 &18.07 &17.68  &13.62 &73.44 &63.31  &15.80 &41.76 &40.85 &40.13 \\\rowcolor{rowgray}
Video &30  &16 &16.86 &16.14  &13.44 &74.39 &55.65  &17.97 &40.88 &42.18 &36.80 \\
Video &20  &16 &13.85 &12.29  &13.58 &73.65 &52.28  &19.13 &42.64 &45.31 &38.95 \\\rowcolor{rowgray}
Video &15  &16 &12.34 &10.36  &13.71 &72.94 &39.20  &25.51 &41.53 &41.76 &41.42 \\
Video &10  &16 &10.83 &8.44   &13.57 &73.67 &35.87  &27.88 &41.76 &40.98 &38.79 \\\rowcolor{rowgray}

\midrule

fNIRS-Wave &24 &16 &7.81  &1.34  &13.48 &74.16 &17.95 &55.72 &45.05 &42.64 &39.42 \\
fNIRS-PSD  &24 &16 &15.05 &13.83 &13.84 &72.25 &46.11 &21.69 &\textbf{52.04} &\textbf{46.59} &\textbf{45.88} \\

\bottomrule
\end{tabular}

\end{table*}
%%%%%%%%%%%%%%%%%%%%%%%%%%%%%%%%%%%%%%%%%%%%%%%%%

%%%%%%%%%%%%%%%%%%%%%%%%%%%%%%%%%%%%%%%%%%%%%%%%%
\begin{table*}
\footnotesize
\caption{Performance and computational/inference cost of the \textit{Segment-Latent} variant with 32 segments.}
\label{table:v3_32_segments}
\centering

\begin{tabular}{P{1.3cm} P{0.50cm} P{0.55cm} P{1.1cm} P{0.9cm} P{2.0cm} P{1.7cm} P{2.0cm} P{1.7cm} P{0.7cm} P{0.6cm} P{0.4cm}}
\toprule

\multirow{2}{*}{\shortstack{Modality}} &
\multirow{2}{*}{\shortstack{\#Chan.}} &
\multirow{2}{*}{\shortstack{\#Segm.}} &
\multicolumn{2}{c}{Computational Cost} &
\multicolumn{4}{c}{Inference Cost} &
\multicolumn{3}{c}{Performance} \\

\cmidrule(lr){4-5}\cmidrule(lr){6-9}\cmidrule(lr){10-12}
& & & Params(M) & GFLOPs &
Latency (ms) GPU$\downarrow$ &
Samples/s GPU$\uparrow$ &
Latency (ms) CPU$\downarrow$ &
Samples/s CPU$\uparrow$ &
Accuracy & Precision & F1 \\

\midrule
\midrule

Video &300 &32 &98.36 &121.43 &90.30 &11.11 &349.35 &2.86  &\underline{46.85} &47.02 &\textbf{45.87} \\\rowcolor{rowgray}
Video &150 &32 &53.09 &63.63  &47.12 &21.22 &179.64 &5.57  &46.34 &45.44 &45.10 \\
Video &100 &32 &37.99 &44.36  &28.17 &35.50 &127.39 &7.85  &44.21 &44.46 &44.23 \\\rowcolor{rowgray}
Video &75  &32 &30.45 &34.73  &21.00 &47.61 &104.65 &9.56  &45.56 &43.20 &42.92 \\
Video &60  &32 &25.92 &28.95  &16.64 &60.09 &98.97  &10.10 &42.59 &42.35 &41.26 \\\rowcolor{rowgray}
Video &50  &32 &22.90 &25.09  &14.15 &70.67 &70.99  &14.09 &\textbf{48.94} &\underline{47.96} &\underline{45.21} \\
Video &43  &32 &20.79 &22.40  &14.12 &70.83 &61.05  &16.38 &41.94 &41.03 &34.14 \\\rowcolor{rowgray}
Video &38  &32 &19.28 &20.47  &13.92 &71.85 &60.03  &16.66 &42.55 &42.14 &41.83 \\
Video &33  &32 &18.07 &18.93  &13.59 &73.59 &66.34  &15.07 &42.50 &42.51 &40.11 \\\rowcolor{rowgray}
Video &30  &32 &16.86 &17.39  &13.96 &71.65 &55.76  &17.93 &41.76 &40.98 &28.89 \\
Video &20  &32 &13.85 &13.53  &14.04 &71.24 &44.17  &22.64 &43.52 &43.97 &41.05 \\\rowcolor{rowgray}
Video &15  &32 &12.34 &11.61  &13.76 &72.70 &40.59  &24.64 &42.78 &\textbf{50.83} &42.32 \\
Video &10  &32 &10.83 &9.68   &13.55 &73.79 &36.66  &27.27 &43.15 &45.97 &32.00 \\\rowcolor{rowgray}

\midrule

fNIRS-Wave &24 &32 &7.81  &2.58  &13.50 &74.06 &22.97 &43.54 &47.18 &43.54 &41.34 \\
fNIRS-PSD  &24 &32 &15.05 &15.07 &14.64 &68.31 &48.88 &20.46 &\textbf{51.16} &\textbf{45.42} &\textbf{45.70} \\

\bottomrule
\end{tabular}

\end{table*}
%%%%%%%%%%%%%%%%%%%%%%%%%%%%%%%%%%%%%%%%%%%%%%%%%

%%%%%%%%%%%%%%%%%%%%%%%%%%%%%%%%%%%%%%%%%%%%%%%%%
\begin{table*}
\footnotesize
\caption{Performance and computational/inference cost of the \textit{Segment-Latent} variant with 64 segments.}
\label{table:v3_64_segments}
\centering

\begin{tabular}{P{1.3cm} P{0.50cm} P{0.55cm} P{1.1cm} P{0.9cm} P{2.0cm} P{1.7cm} P{2.0cm} P{1.7cm} P{0.7cm} P{0.6cm} P{0.4cm}}
\toprule

\multirow{2}{*}{\shortstack{Modality}} &
\multirow{2}{*}{\shortstack{\#Chan.}} &
\multirow{2}{*}{\shortstack{\#Segm.}} &
\multicolumn{2}{c}{Computational Cost} &
\multicolumn{4}{c}{Inference Cost} &
\multicolumn{3}{c}{Performance} \\

\cmidrule(lr){4-5}\cmidrule(lr){6-9}\cmidrule(lr){10-12}
& & & Params(M) & GFLOPs &
Latency (ms) GPU$\downarrow$ &
Samples/s GPU$\uparrow$ &
Latency (ms) CPU$\downarrow$ &
Samples/s CPU$\uparrow$ &
Accuracy & Precision & F1 \\

\midrule
\midrule

Video &300 &64 &98.36 &123.92 &96.29 &10.39 &364.32 &2.74  &45.60 &45.69 &\underline{44.77} \\\rowcolor{rowgray}
Video &150 &64 &53.09 &66.11  &45.81 &21.83 &199.34 &5.02  &41.02 &39.78 &38.79 \\
Video &100 &64 &37.99 &46.85  &27.71 &36.09 &140.76 &7.10  &\textbf{46.57} &\underline{47.04} &39.71 \\\rowcolor{rowgray}
Video &75  &64 &30.45 &37.21  &21.33 &46.88 &104.22 &9.60  &\underline{45.79} &44.85 &44.30 \\
Video &60  &64 &25.92 &31.43  &16.50 &60.59 &90.27  &11.08 &43.15 &43.16 &43.08 \\\rowcolor{rowgray}
Video &50  &64 &22.90 &27.58  &13.56 &73.74 &72.85  &13.73 &43.47 &46.56 &42.78 \\
Video &43  &64 &20.79 &24.88  &13.63 &73.36 &67.00  &14.93 &42.36 &42.31 &41.77 \\\rowcolor{rowgray}
Video &38  &64 &19.28 &22.95  &13.98 &71.51 &60.98  &16.40 &43.38 &43.04 &42.90 \\
Video &33  &64 &18.07 &21.41  &13.69 &73.04 &69.37  &14.42 &44.91 &44.07 &43.94 \\\rowcolor{rowgray}
Video &30  &64 &16.86 &19.87  &14.23 &70.25 &55.45  &18.03 &41.11 &42.31 &33.15 \\
Video &20  &64 &13.85 &16.02  &13.76 &72.68 &47.68  &20.97 &41.02 &\textbf{50.94} &\textbf{49.59} \\\rowcolor{rowgray}
Video &15  &64 &12.34 &14.09  &13.52 &73.95 &49.97  &20.01 &45.32 &45.30 &44.47 \\
Video &10  &64 &10.83 &12.16  &13.72 &72.87 &41.28  &24.23 &41.25 &40.57 &40.19 \\\rowcolor{rowgray}

\midrule

fNIRS-Wave &24 &64 &7.81  &5.07  &14.09 &70.99 &31.94 &31.31 &46.85 &\textbf{42.19} &\textbf{41.81} \\
fNIRS-PSD  &24 &64 &15.05 &17.56 &13.99 &71.48 &51.83 &19.29 &\textbf{49.81} &29.01 &36.67 \\

\bottomrule
\end{tabular}

\end{table*}
%%%%%%%%%%%%%%%%%%%%%%%%%%%%%%%%%%%%%%%%%%%%%%%%%

%%%%%%%%%%%%%%%%%%%%%%%%%%%%%%%%%%%%%%%%%%%%%%%%%

%%%%%%%%%%%%%%%%%%%%%%%%%%%%%%%%%%%%%%%%%%%%%%%%

\begin{table*}
\footnotesize
\caption{Performance and computational/inference costs for modality fusion.}
\label{table:fusion}
\centering
\renewcommand{\arraystretch}{1.15}

\begin{tabular}{
P{1.3cm}
P{0.40cm}
P{0.45cm}
P{0.45cm}
P{0.85cm}
P{0.7cm}
P{1.95cm}
P{1.65cm}
P{1.95cm}
P{1.65cm}
P{0.65cm}
P{0.55cm}
P{0.5cm}
}
\toprule

\multirow{2}{*}{\shortstack{Modality}} &
\multirow{2}{*}{\shortstack{\#Chan.}} &
\multirow{2}{*}{\shortstack{\#Segm.}} &
\multirow{2}{*}{\shortstack{Fusion}} &
\multicolumn{2}{c}{Computational Cost} &
\multicolumn{4}{c}{Inference Cost} &
\multicolumn{3}{c}{Performance} \\

\cmidrule(lr){5-6}
\cmidrule(lr){7-10}
\cmidrule(lr){11-13}

& & & &
Params(M) &
GFLOPs &
Latency (ms) GPU$\downarrow$ &
Samples/s GPU$\uparrow$ &
Latency (ms) CPU$\downarrow$ &
Samples/s CPU$\uparrow$ &
Accuracy &
Precision &
F1 \\

\midrule
\midrule

\shortstack{fNIRS-Wave\\fNIRS-PSD} &
\shortstack{24\\24} &
32 &
stack &
15.15 &
15.20 &
13.79 &
72.54 &
49.00 &
20.41 &
48.52 &
\underline{45.62} &
44.62 \\

\multicolumn{13}{@{}c@{}}{%
\cellcolor{rowgray}%
\begin{tabular}{
P{1.3cm}
P{0.40cm}
P{0.45cm}
P{0.45cm}
P{0.85cm}
P{0.7cm}
P{1.95cm}
P{1.65cm}
P{1.95cm}
P{1.65cm}
P{0.65cm}
P{0.55cm}
P{0.5cm}
}
\shortstack{Video\\fNIRS-Wave} &
\shortstack{20\\24} &
32 &
stack &
13.95 &
13.66 &
13.57 &
73.69 &
53.00 &
18.87 &
47.96 &
\textbf{46.00} &
\underline{45.19}
\end{tabular}%
} \\

\shortstack{Video\\fNIRS-PSD} &
\shortstack{20\\24} &
32 &
stack &
21.09 &
22.78 &
14.03 &
71.30 &
61.53 &
16.25 &
\textbf{50.93} &
45.59 &
\textbf{45.37} \\

\multicolumn{13}{@{}c@{}}{%
\cellcolor{rowgray}%
\begin{tabular}{
P{1.3cm}
P{0.40cm}
P{0.45cm}
P{0.45cm}
P{0.85cm}
P{0.7cm}
P{1.95cm}
P{1.65cm}
P{1.95cm}
P{1.65cm}
P{0.65cm}
P{0.55cm}
P{0.5cm}
}
\shortstack{Video\\fNIRS-Wave\\fNIRS-PSD} &
\shortstack{20\\24\\24} &
32 &
stack &
21.19 &
22.91 &
14.74 &
67.84 &
68.89 &
14.52 &
\underline{49.63} &
32.49 &
38.68
\end{tabular}%
} \\

\bottomrule
\end{tabular}

\end{table*}

%%%%%%%%%%%%%%%%%%%%%%%%%%%%%%%%%%%%%%%%%%%%%%%%

\subsection{Modality Fusion}
\label{sec:modality_fusion}
Regarding the fusion of modalities, a simple method, \textit{stack}, is employed that concatenates modality representations along the channel dimension, enabling a single shared backbone to process both unimodal and multimodal inputs.
For the fusion approach, $S=32$ segments were selected, as they offer strong performance with balanced computational and inference costs. Table \ref{table:fusion} reports the corresponding fusion results.
For fNIRS-only fusion, combining waveform and PSD improves performance over individual representations, achieving $48.52\%$ accuracy, $45.62\%$ precision, and $44.62\%$ F1. 
For video--fNIRS fusion, stacking video with the waveform representation achieves $47.96\%$ accuracy and the highest precision among these settings ($46.00\%$), while stacking video with PSD provides the strongest overall multimodal accuracy at $50.93\%$ and the highest F1 score of $45.37\%$. In contrast, stacking video, waveform, and PSD together increases computational cost without improving performance, with precision and F1 dropping to $32.49\%$ and $38.68\%$, respectively.
Regarding the waveform representation and its fusion with either video or PSD inputs, the original 1D signal is first projected onto a 2D grid via interpolation and subsequently stacked with the corresponding tensor along the channel dimension:
\begin{equation}
\mathbf{X}^{\text{stack}}=\mathrm{Cat}_{C}\!\big(\mathbf{X}^{\text{vid}},\ \mathcal{I}(\mathbf{X}^{\text{wave}})\big),
\end{equation}
where $\mathcal{I}$ denotes the interpolation from the $C\times L$ waveform grid to a single $H\times W$ tensor.

\section{Comparison with Existing Methods}
\label{comparison_testing}

For the final test-set evaluation, the video--fNIRS waveform stack fusion was selected over the video--PSD combination, as it achieves a more favorable accuracy--efficiency trade-off on the validation set, with substantially lower computational cost ($13.95$M parameters, $13.66$ GFLOPs versus $21.09$M and $22.78$ GFLOPs), better aligning with the real-time deployment objective of the proposed framework.

Table \ref{table:ai4pain_test} reports a direct comparison on the \textit{AI4Pain} testing set with video-, fNIRS-, and multimodal-based approaches. 
The handcrafted-feature SVM framework in \cite{ai4pain_2024} reports $43.30\%$ using fNIRS alone, $40.10\%$ with video, and $41.70\%$ when both modalities are combined, highlighting the limited capacity of classical pipelines to exploit cross-modal information. Deep learning models show improved performance, with the transformer-based model \cite{gkikas_tsiknakis_painvit_2024} reaching $46.67\%$ using multimodal inputs. At the same time, CNN and hybrid architectures achieve performance gains of $49.00\%$ for video-only \cite{prajod_schiller_2024} and $51.33\%$ for multimodal fusion \cite{vianto_2025}.
Physiological-signal--focused methods also demonstrate strong results, with the ensemble classifier of \cite{khan_aziz_2025} achieving $53.66\%$ using fNIRS features, and the transformer-based video model of \cite{nguyen_yang_2024} reporting 
$55.00\%$ accuracy. The multimodal framework \cite{gkikas_rojas_painformer_2025} achieved $55.69\%$ for joint video--fNIRS modeling.
The proposed framework achieves high performance in both unimodal and multimodal settings. Using fNIRS alone attains $53.67\%$ accuracy, comparable to the best physiological-only method, while video-only input reaches $57.00\%$, surpassing existing video-based approaches. 
The stacked video--fNIRS fusion achieves the highest reported performance of $57.33\%$, demonstrating that the unified tokenization framework effectively integrates heterogeneous modalities within a single shared model.

%%%%%%%%%%%%%%%%%%%%%%%%%%%%%%%%%%%%%%%%%%%%%%%%
\begin{table}
\footnotesize
\caption{Comparison of studies on the \textit{AI4Pain} testing set.}
\label{table:ai4pain_test}
\centering
\renewcommand{\arraystretch}{1.15}

\begin{tabular}{
P{0.6cm}
P{0.7cm}
P{0.7cm}
P{1.5cm}
P{1.9cm}
P{0.9cm}
}
\toprule

\multirow{2}{*}{\shortstack{Study}} &
\multicolumn{2}{c}{Modality} &
\multicolumn{2}{c}{Method} &
\multirow{2}{*}{\shortstack{Accuracy}} \\

\cmidrule(lr){2-3}
\cmidrule(lr){4-5}

& Video & fNIRS & Features & Model & \\

\midrule
\midrule

& \xmark & \checkmark & & & 43.30 \\
& \checkmark & \xmark & & & 40.10 \\
\multirow{-3}{*}{\cite{ai4pain_2024}}
& \checkmark & \checkmark &
\multirow{-3}{*}{Handcrafted} &
\multirow{-3}{*}{SVM} &
41.70 \\

% Gray row: [27]
\multicolumn{6}{@{}c@{}}{%
\cellcolor{rowgray}%
\begin{tabular}{
P{0.6cm}
P{0.7cm}
P{0.7cm}
P{1.5cm}
P{1.9cm}
P{0.9cm}
}
\cite{gkikas_tsiknakis_painvit_2024} &
\checkmark &
\checkmark &
Deep &
Transformer &
46.67
\end{tabular}%
} \\

\cite{prajod_schiller_2024}
& \checkmark & \xmark
& Deep & 2D CNN & 49.00 \\

% Gray row: [66]
\multicolumn{6}{@{}c@{}}{%
\cellcolor{rowgray}%
\begin{tabular}{
P{0.6cm}
P{0.7cm}
P{0.7cm}
P{1.5cm}
P{1.9cm}
P{0.9cm}
}
\cite{vianto_2025} &
\checkmark &
\checkmark &
Deep &
CNN-Transformer &
51.33
\end{tabular}%
} \\

\cite{khan_aziz_2025}
& \xmark & \checkmark
& Handcrafted & ENS & 53.66 \\

% Gray row: [53]
\multicolumn{6}{@{}c@{}}{%
\cellcolor{rowgray}%
\begin{tabular}{
P{0.6cm}
P{0.7cm}
P{0.7cm}
P{1.5cm}
P{1.9cm}
P{0.9cm}
}
\cite{nguyen_yang_2024} &
\checkmark &
\xmark &
Deep &
Transformer &
55.00
\end{tabular}%
} \\

& \xmark & \checkmark & & & 52.60 \\
& \checkmark & \xmark & & & 53.67 \\
\multirow{-3}{*}{\cite{gkikas_rojas_painformer_2025}}
& \checkmark & \checkmark &
\multirow{-3}{*}{Deep} &
\multirow{-3}{*}{Transformer} &
55.69 \\

\midrule

% Continuous gray block: Our
\multicolumn{6}{@{}c@{}}{%
\cellcolor{rowgray}%
\begin{tabular}{
>{\centering\arraybackslash}m{0.6cm}
>{\centering\arraybackslash}m{0.7cm}
>{\centering\arraybackslash}m{0.7cm}
>{\centering\arraybackslash}m{1.5cm}
>{\centering\arraybackslash}m{1.9cm}
>{\centering\arraybackslash}m{0.9cm}
}
Our &
\shortstack{\xmark\\\checkmark\\\checkmark} &
\shortstack{\checkmark\\\xmark\\\checkmark} &
Deep &
Transformer &
\shortstack{
53.67$^\dagger$\\
57.00$^\ddagger$\\
\textbf{57.33}\textsuperscript{\ding{93}}
}
\end{tabular}%
} \\

\bottomrule
\end{tabular}

\vspace{2pt}

\parbox{\columnwidth}{%
\scriptsize
\checkmark\ indicates that the modality is used;
\xmark\ indicates that it is not used.
ENS: Ensemble Classifier.
$\dagger$: PSD.
$\ddagger$: 20 channels.
\ding{93}: Video and fNIRS-Wave stack fusion.
}

\end{table}

\section{Conclusion}
This work introduced a unified tokenization framework for pain recognition that combines heterogeneous 3D modalities within a single shared backbone, including facial videos, raw fNIRS waveforms, and spectrogram-based fNIRS representations. This approach preserves modality structure through axis folding, Fourier-augmented tokenization, and segment-based latent aggregation, avoiding modality-specific architectures while maintaining efficient inference.
Experiments on the \textit{AI4Pain} benchmark dataset demonstrated high performance results across unimodal and multimodal settings. The \textit{Segment-Latent} variant provided a strong balance between accuracy and computational cost, while simple \textit{stack} fusion enabled effective video--fNIRS integration without additional fusion modules. 
The proposed framework achieved state-of-the-art performance on the test set. At the same time, the inference-cost metrics demonstrate real-time performance on both GPU and CPU hardware, which is necessary for practical deployment.

\section*{Safe and Responsible Innovation Statement}
All experiments were conducted on the \textit{AI4Pain} dataset, released by the challenge organizers. Participants were free from neurological or psychiatric disorders, chronic pain, and regular medication use. Informed consent was obtained after full disclosure of the experimental procedures, and the study protocol was approved by the Human Ethics Committee of the University of Canberra (\textit{approval number: 11837}). 
The proposed framework aims to enable continuous, objective pain assessment, reducing dependence on subjective clinical evaluation. Real-world deployment would require careful validation through dedicated trials before integration into clinical practice. 
As the dataset was collected under controlled conditions and represents a specific demographic group, its ability to generalize to diverse populations, age groups, and varying cultural expressions of pain is limited; further research is required.

\bibliographystyle{ACM-Reference-Format}
\bibliography{library}

\end{document}